\documentclass[10pt,journal, compsoc]{IEEEtran}
\usepackage{amsmath,amsfonts}
\usepackage{algorithm}
\usepackage{algorithmic}
\usepackage{array}
\usepackage[caption=false,font=normalsize,labelfont=sf,textfont=sf]{subfig}
\usepackage{textcomp}
\usepackage{stfloats}
\usepackage{url}
\usepackage{verbatim}
\usepackage{graphicx}
\def\BibTeX{{\rm B\kern-.05em{\sc i\kern-.025em b}\kern-.08em
    T\kern-.1667em\lower.7ex\hbox{E}\kern-.125emX}}
\usepackage{balance}

\usepackage{cite}
\usepackage{amsthm}
\usepackage{epstopdf}
\usepackage{multirow}
\usepackage{booktabs}
\usepackage{diagbox}
\usepackage{bm}

\newtheorem{theorem}{Theorem}
\newtheorem{lemma}{Lemma}

\begin{document}
\title{Causally-Aware Unsupervised Feature Selection Learning}
\author{Zongxin Shen, Yanyong Huang, Dongjie Wang, Minbo Ma, Fengmao~Lv, Tianrui Li,~\IEEEmembership{Senior Member,~IEEE}

\thanks{Zongxin~Shen and Yanyong~Huang are with the Joint Laboratory of Data Science and Business Intelligence, School of  Statistics,  Southwestern University of Finance and Economics, Chengdu 611130, China (e-mail: zxshen@smail.swufe.edu.cn; huangyy@swufe.edu.cn);}
\thanks{Dongjie Wang is with the Department of Electrical Engineering and Computer Science, University of Kansas, Lawrence, KS 66045, USA (e-mail: wangdongjie@ku.edu);}
\thanks{ Minbo Ma, Fengmao~Lv, and Tianrui Li are with the School of Computing and Artificial Intelligence, Southwest Jiaotong University, Chengdu 611756, China (e-mail: minbo46.ma@gmail.com; fengmaolv@126.com; trli@swjtu.edu.cn).}}

\markboth{Journal of \LaTeX\ Class Files,~Vol.~18, No.~9, September~2020}%
{How to Use the IEEEtran \LaTeX \ Templates}

\maketitle

\begin{abstract}
    Unsupervised feature selection (UFS) has recently gained attention for its effectiveness in processing unlabeled high-dimensional data. However, existing methods overlook the intrinsic causal mechanisms within the data, resulting in the selection of irrelevant features and poor interpretability. Additionally, previous graph-based methods fail to account for the differing impacts of non-causal and causal features in constructing the similarity graph, which leads to false links in the generated graph. To address these issues, a novel UFS method, called Causally-Aware UnSupErvised Feature Selection learning (CAUSE-FS), is proposed. CAUSE-FS introduces a novel causal regularizer that reweights samples to balance the confounding distribution of each treatment feature. This regularizer is subsequently integrated into a generalized unsupervised spectral regression model to mitigate spurious associations between features and clustering labels, thus achieving causal feature selection. Furthermore, CAUSE-FS employs causality-guided hierarchical clustering to partition features with varying causal contributions into multiple granularities. By integrating similarity graphs learned adaptively at different granularities, CAUSE-FS increases the importance of causal features when constructing the fused similarity graph to capture the reliable local structure of data. Extensive experimental results demonstrate the superiority of CAUSE-FS over state-of-the-art methods, with its interpretability further validated through feature visualization.
\end{abstract}


\section{Introduction}
\IEEEPARstart{H}{igh-dimensional} often contain numerous redundant and noisy features, which can result in the curse of dimensionality and degrade the performance of downstream tasks~\cite{KaiLagemann2023,ahadzadeh2023sfe,islam2024deciphering}. Feature selection has proven to be an effective technique for reducing the dimensionality of high-dimensional data~\cite{li2017feature,hou2023adaptive}. Since obtaining large amounts of labeled data in practice is laborious and time-consuming, unsupervised feature selection (UFS), which identifies informative features from unlabeled high-dimensional data, has garnered increasing attention in recent years~\cite{solorio2020review,VCSDFS,li2024exploring}. 

\begin{figure}[t]
	\centering
	\includegraphics[width=\columnwidth]{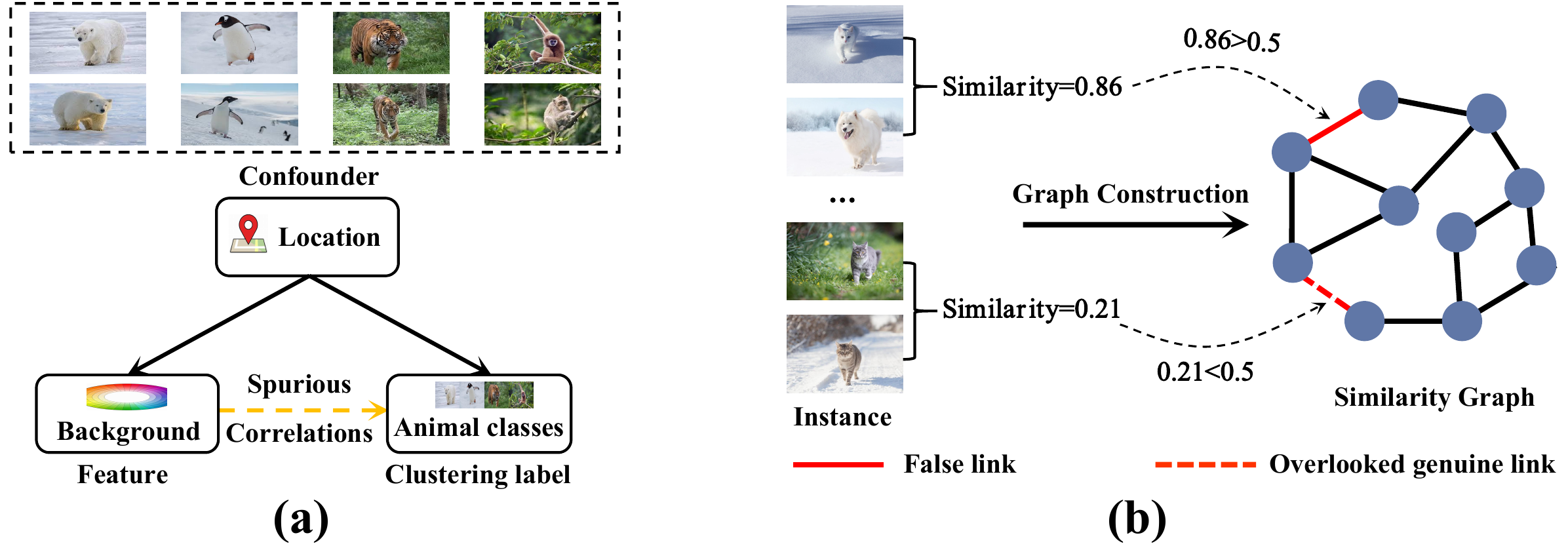} 
	\caption{The examples of spurious correlation between features and clustering labels caused by confounders (left), and the unreliable links in the similarity graph result from ignoring the importance of different features (right).}
	\label{illustration1}
\end{figure}

However, existing UFS methods emphasize the discriminative power of features, neglecting the underlying causality within data. These methods typically identify the most discriminative features by assessing the dependence between features and inferred clustering labels in the unsupervised scenario~\cite{zhu2017subspace,lin2021unsupervised,FSDK}. Confounding factors that simultaneously influence both features and clustering labels can induce spurious correlations between them. As a result, conventional UFS methods often select numerous irrelevant features and lack interpretability. As an example of animal clustering, as shown in Fig.~\ref{illustration1} (a), photos of polar bears and penguins are often captured against white backgrounds (e.g., polar regions), while tigers and monkeys are typically photographed against green backgrounds (e.g., forests). In this context, location acts as a confounder, influencing both the clustering labels (animal classes) and the background color. This confounder creates a backdoor path between the background and the clustering labels, resulting in spurious correlations. By ignoring confounding effects, existing UFS methods rely on spurious correlations for feature selection, leading to the inclusion of irrelevant features, such as the aforementioned background features. Since background features are ineffective for distinguishing animals classes, their inclusion degrades the performance of downstream clustering tasks. Besides, because spurious correlations fail to capture the true causal relationships between features and clustering labels, the selected features lack interpretability. Although several causality-based feature selection methods have been proposed recently, they are impractical in unsupervised scenarios due to their reliance on label information~\cite{yu2022causal,quinzan2023drcfs,yu2021unified}.
\begin{figure}[t]
	\centering
	\includegraphics[width=\columnwidth]{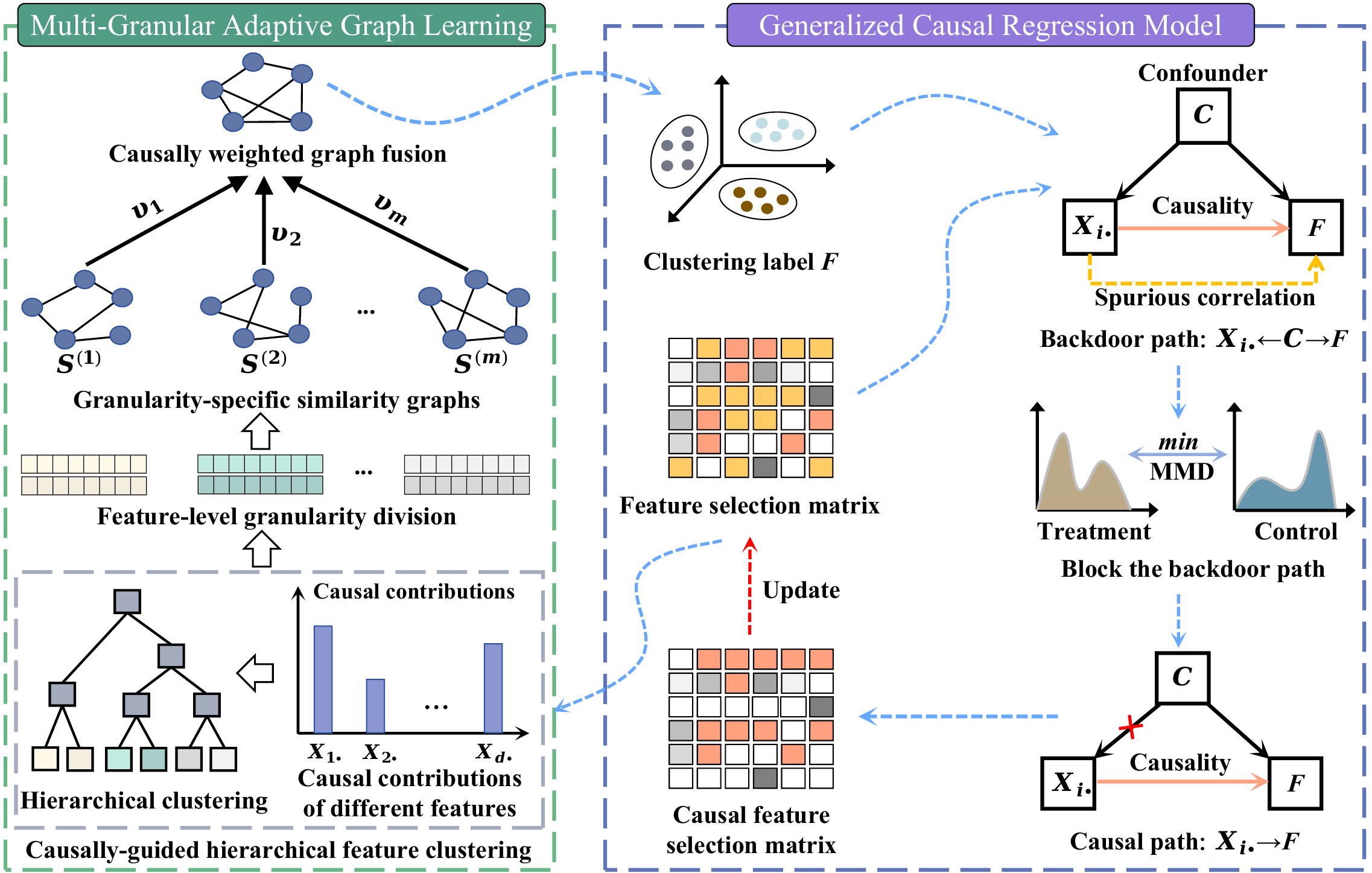} 
	\caption{The framework of the proposed CAUFS-FS.}
	\label{framework}
\end{figure}

Furthermore, previous studies have demonstrated that preserving the local manifold structure of data through the construction of various graphs is crucial for UFS~\cite{FSDK,HSL,li2021unsupervised}. However, conventional graph-based UFS methods overlook the differing impacts of causal and non-causal features during the construction of similarity graphs, resulting in the generation of many unreliable links. For instance, as shown in Fig.~\ref{illustration1} (b), due to the failure to prioritize causal features (such as the nose and ear) over non-causal features (like the background) in constructing a similarity graph, a false link between two different categories (cats and dogs) would be created owing to the high similarity of snowy backgrounds. Similarly, the significant difference between grassy and snowy backgrounds leads to the absence of intra-class links among cats. Unreliable links in the constructed similarity graph will degrade the performance of existing UFS methods.

To address the aforementioned issues, we propose a novel UFS method, named Causally-Aware UnSupErvised Feature Selection learning (CAUSE-FS). Specifically, we first integrate the feature selection into an unsupervised spectral regression model equipped with a novel causal regularizer. Benefiting from this causal regularizer, the proposed model effectively mitigates spurious correlations between features and clustering labels by balancing the confounding distribution of each treatment feature, thereby identifying causally informative features. Then, we employ causally-guided hierarchical clustering to partition features with varying causal contributions into multiple granularities. By adaptively learning similarity graphs at each granularity, we facilitate their weighted fusion by leveraging the causal contributions of features at different granularities as weights. In this way, CAUSE-FS can emphasize causal features in the construction of the similarity graph while attenuating the influence of non-causal features. Therefore, CAUSE-FS can capture reliable connections among instances within the resulting graph, thereby preserving the local structure of data more accurately. Fig.~\ref{framework} illustrates the framework of CAUSE-FS. The main contributions of this paper are summarized as follows:

\begin{enumerate}
\item[\textbullet] To the best of our knowledge, the proposed CAUSE-FS is the first work for causal feature selection in unsupervised scenarios. CAUSE-FS integrates feature selection and confounder balancing into a unified learning framework, effectively mitigating spurious correlations between features and clustering labels to identify causally informative features.

\item[\textbullet] We proposed a novel multi-granular adaptive similarity graph learning method by integrating the causal contributions of features. The learned similarity graph can more reliably capture relationships between samples by assigning higher weights to causal features and lower weights to non-causal features.

\item[\textbullet] An efficient alternative optimization algorithm is developed to solve the proposed CAUSE-FS, and comprehensive experiments demonstrate the superiority of CAUSE-FS over other state-of-the-art (SOTA) methods. Furthermore, its interpretability is validated through feature visualization.
\end{enumerate}

The rest of this paper is organized as follows. Section 2 offers a brief review of related work on UFS. Section 3 provides a detailed introduction to the proposed method CAUSE-FS. Section 4 outlines an effective solution for the proposed method, followed by convergence and complexity analysis in Section 5. Section 6 presents extensive experiments and a comparative analysis of the results. Finally, Section 7 concludes the paper.

\section{Related Work}\label{sec:Related work}

In this section, we briefly review some recent UFS methods. CNAFS integrates self-expression and pseudo-label learning into a unified optimization framework while incorporating manifold regularization to preserve the local geometric structure of data~\cite{CNAFS}. Nie et al. propose to simultaneously perform feature selection and similarity graph learning, and constrain the similarity matrix to include exactly $k$ nearest neighbors for local structure preservation~\cite{nie2019structured}. DUFS leverages information theory to calculate pairwise dependencies between features and integrates this dependency-based approach into a regularized regression model to identify discriminative and non-redundant feature subsets~\cite{DUFS}. VCSDFS proposes a novel variance-covariance distance to capture the statistical correlations among features and combines this distance with subspace learning to select features~\cite{VCSDFS}. Based on feature-level and clustering-level analysis, Zhou~et al. propose a bi-level ensemble feature selection method~\cite{BLFSE}. FSDK learns pseudo-labels to guide the least-squares regression learning and imposes the $\ell_{2,p}$-norm regularizer on the feature selection matrix to satisfy different sparsity requirements~\cite{FSDK}. OEDFS proposes an autoencoder-based method for discriminative feature selection by combining self-representation learning and non-negative matrix factorization~\cite{OEDFS}. HSL adaptively learns an optimal graph Laplacian matrix by incorporating first-order and high-order similarities, thereby effectively preserving the geometric structure of data for feature selection~\cite{HSL}. Chen et al. employ nonnegative tensor CP decomposition to preserve the multi-dimensional structure of data during the selection process~\cite{chen2022unsupervised}. FDNFS constructs a feature graph to capture feature-level correlations and embeds this feature graph into a sparse regression model to identify important features. Moreover, FDNFS applies manifold regularization to both the pseudo-label matrix and the reconstruction coefficient matrix, thereby exploiting the intrinsic structure of high-dimensional data to improve feature selection performance~\cite{FDNFS}. However, as discussed earlier, existing UFS methods suffer from two significant  limitations: (\romannumeral1) They concentrate solely on identifying features that maximize discriminative performance, neglecting the underlying causal mechanisms that could confound the feature selection process. (\romannumeral2) They fail to consider the importance of causal and non-causal features in constructing the similarity graph, resulting in unreliable graphs for local structure preservation and suboptimal feature selection performance.

\section{Causally-Aware Unsupervised Feature Selection Learning}\label{sec:proposed method}
In this section, we first summarize the notations and definitions used in this paper. We then introduce the proposed method CAUSE-FS, which consists of two modules. The first module uses a generalized causal regression model to address confounding bias between features and clustering labels, aiming to learn causally informative features and clustering labels.  The second module incorporates the causal contributions of features into adaptive similarity graph learning to more effectively capture the local data structure and enhance feature selection performance.

\subsection{Notations}
For any matrix $\bm{A}\in \mathbb{R}^{p \times q}$, the $(i,j)$-th entry, the $i$-th row and the $j$-th column are denoted by $\mathnormal{A}_{ij}$, $\bm{A}_{i \cdot}$, and $\bm{A}_{\cdot j}$, respectively. The transpose of matrix $\bm{A}$ is denoted by $\bm{A}^{\top}$, and the trace of $\bm{A}$ is defined as $\operatorname{Tr}(\bm{A})$. $\|\bm{A}\|_{\mathrm{F}}=\sqrt{\sum_{i=1}^{p}\sum_{j=1}^{q}\mathnormal{A}_{ij}^{2}}$ is the Frobenius norm, and $\|\bm{A}\|_{2,1}=\sum_{i=1}^{p}\sqrt{\sum_{j=1}^{q}\mathnormal{A}_{ij}^{2}}$ denotes the $\ell_{2,1}$-norm. $\bm{I}$ represents the identity matrix, and $\bm{1}=[1,\dots,1]^{\top}$ stands for a column vector with all entries being one. 

\subsection{Generalized Causal Regression Model for UFS} In order to identify the causally informative features, we propose to learn the causal contribution of each feature to differentiating clusters. To this end, we first integrate feature selection into an unsupervised spectral regression model. It can identify the important features by exploiting the dependence between features and clustering labels. The corresponding objective function is presented as follows:
\begin{equation}\label{3.1}
\begin{aligned}
&\min_{\bm{W},\bm{F}}\alpha\|\bm{X}^{\top}\bm{W} \!-\! \bm{F} \|_{\mathrm{F}}^{2}+\operatorname{Tr}( \bm{F}^{\top}\bm{L}_{S}\bm{F})+\lambda \|\bm{W}\|_{2,1}\\
&\text { s.t. } \bm{F}^{\top}\bm{F}=\bm{I},
\end{aligned}
\end{equation}
where $\bm{X} \in \mathbb{R}^{d \times n}$ denotes the data matrix, with $n$ and $d$ respectively representing the number of samples and feature dimensions, $\bm{L}_{S}$ is the Laplacian matrix of similarity matrix $\bm{S} \in \mathbb{R}^{n \times n}$. $\bm{F} \in \mathbb{R}^{n \times h}$ denotes the clustering label, and $\bm{W} \in \mathbb{R}^{d \times h}$ is the feature selection matrix, with $h$ denoting the dimension of the low-dimensional subspace. For $\bm{W}$, the $i$-th row $\bm{W}_{i \cdot}$ reflects the dependence between the feature $\bm{X}_{i \cdot}$ and the clustering label $\bm{F}$. A stronger dependence indicates a greater contribution of $\bm{X}_{i \cdot}$ to cluster differentiation. Thus, $\|\bm{W}_{i \cdot}\|_{2}^{2}$ can measure the importance of the $i$-th feature. Additionally, the row-sparsity of $\bm{W}$ enforced by $\ell_{2,1}$-norm regularization facilitates the removal of less significant features during feature selection.

As previously discussed, each row in  $\bm{W}$ may exhibit a spurious correlation between the feature and the clustering label, attributed to confounding factors. To address this issue, we propose a novel causal regularizer to reduce confounding effects. Specifically, for a given treatment feature $\bm{X}_{i\cdot}$, the remaining features obtained by excluding those unrelated to either the clustering labels or $\bm{X}_{i\cdot}$ from the full feature set are denoted as $\bm{Z}^{i}$ and treated as confounders. A stepwise strategy for identifying these confounding features will be detailed below. Then, the samples with the features $\bm{Z}^{i}$ are divided into treatment and control groups based on the values of the treatment feature. Without loss of generality, we assume that the treatment feature values are binary, taking on values of 0 or 1, where samples with a value of 1 are assigned to the treatment group and those with a value of 0 are assigned to the control group. To capture the true causality, it is essential to block the backdoor path by aligning the confounder distribution between the treatment and control groups. Thus, we sequentially treat each feature as a treatment feature and compute global sample weights to achieve a balanced confounder distribution for each respective feature. It can be expressed as follows:
\begin{equation}\label{3.2}
\begin{aligned}
&\min_{\bm{\mu}} \sum_{r=1}^{d}\| \frac{1}{|\Lambda_{t}|}\sum_{i\in\Lambda_{t}}\phi(\mu_{i}Z_{\cdot i}^{r}) - \frac{1}{|\Lambda_{c}|}\sum_{j\in\Lambda_{c}}\phi(\mu_{j}Z_{\cdot j}^{r})\|_{\mathcal{H}_{\kappa}}^{2}\\
&\text { s.t. } \mu_{i} \geq 0, \bm{1}^{\top}\bm{\mu}=1,
\end{aligned}
\end{equation}
where $\bm{\mu} \in \mathbb{R}^{n \times 1}$ is the sample weights vector, $\Lambda_{t}$ and $\Lambda_{c}$ respectively represent the set of sample indexes for the treatment and control groups, and $|\Lambda_t|$ denotes the cardinality of $\Lambda_t$. In Eq. (\ref{3.2}), Maximum Mean Discrepancy~\cite{gretton2006kernel} is used to measure distribution differences, which can capture complex nonlinear relationships by embedding distributions into the Reproducing Kernel Hilbert Space (RKHS). Here, $\phi(\mu_{i}Z_{\cdot i}^{r})=\kappa(\mu_{i}Z_{\cdot i}^{r},\cdot)$ denotes the feature map of $\mu_{i}Z_{\cdot i}^{r}$ with kernel function $\kappa(\cdot,\cdot)$, and $\| \cdot\|_{\mathcal{H}_{\kappa}}$ is the RKHS norm~\cite{cui2020calibrated}. 

By combining Eqs. (\ref{3.1}) and (\ref{3.2}), we can obtain the proposed generalized causal regression  model:
\begin{equation}\label{3.3}
\begin{aligned} 
\min_{\bm{W}\!,\bm{F}\!,\bm{\mu}}& \sum_{i=1}^{n}\!\alpha\|\mu_{i}\bm{X}_{\cdot i}^{\top}\bm{W} \!\!-\! \bm{F}_{i \cdot} \|_{\mathrm{2}}^{2}\!+\!\operatorname{Tr}( \bm{F}^{\top}\bm{L}_{S}\bm{F})\!+\!\lambda \|\bm{W}\|_{2,1}\\
&+\!\beta \sum_{r=1}^{d}\| \frac{1}{|\Lambda_{t}|}\!\sum_{i\in\Lambda_{t}}\!\phi(\mu_{i}Z_{\cdot i}^{r}) \!-\! \frac{1}{|\Lambda_{c}|}\!\sum_{j\in\Lambda_{c}}\!\phi(\mu_{j}Z_{\cdot j}^{r})\|_{\mathcal{H}_{\kappa}}^{2}\\
\text { s.t. } \bm{F}&^{\top}\bm{F}=\bm{I}, \mu_{i} \geq 0, \bm{1}^{\top}\bm{\mu}=1.
\end{aligned}
\end{equation}
By jointly learning feature selection and confounder balancing, each row of the feature selection matrix $\bm{W}$ in Eq. (\ref{3.3}) can capture the causal dependence between features and clustering labels after reducing confounding bias~\cite{FSDK,HSL}.

In Eq.~(\ref{3.3}), due to the absence of prior knowledge about the true confounding factors, we propose a stepwise strategy to identify confounding features for each treatment feature. This strategy involves removing irrelevant variables that do not influence the clustering labels, as well as adjustment variables that affect the clustering labels but lack a causal relationship with the treatment feature. The specifics of this strategy are provided  below. For a given treatment feature $\bm{X}_{i\cdot}$, we construct an initial confounder set $\bm{Z}^{i}=\bm{X} \backslash \bm{X}_{i\cdot}$ by excluding $\bm{X}_{i\cdot}$ from the original feature set.  Subsequently, using the feature selection matrix $\bm{W}$ derived from Eq. (\ref{3.3}), we progressively eliminate irrelevant and adjustment variables through the following steps: (\romannumeral1) Removing irrelevant variables: Features corresponding to rows in the matrix $\bm{W}$ with all-zero elements indicate no causal dependence with the clustering labels. These features are treated as irrelevant variables, denoted as $\bm{U}$, and are removed from the initial confounder set, resulting in $\bm{Z}^{i}=\bm{X} \backslash \bm{X}_{i\cdot} \backslash \bm{U}$. (\romannumeral2) Removing adjustment variables: Based on the causal contributions represented in matrix $\bm{W}$, we calculate the correlation between each feature in $\bm{Z}^{i}$ and the treatment feature. These correlations are normalized to a range of 0 to 1. Features with correlations below the threshold $\sigma$ ($\sigma\leq 0.2$) are denoted as adjustment variables $\bm{E}^{i}$ and removed from the confounder set, resulting in  $\bm{Z}^{i}=\bm{X} \backslash \bm{X}_{i\cdot} \backslash \bm{U} \backslash \bm{E}^{i}$. By using the stepwise strategy described above, we can progressively refine the confounder set, mitigating the influence of irrelevant and adjustment variables on confounder balancing. As a result, the learned feature selection matrix $\bm{W}$ more reliably captures the causal relationships between features and clustering labels.
\subsection{Multi-Granular Adaptive Graph Learning} 
Previous studies have shown that similarity-induced graphs, which preserve the local structure of data, are effective in enhancing the performance of unsupervised feature selection. However, conventional graph-based UFS methods fail to account for the varying impacts of causal and non-causal features during the construction of similarity-induced graphs, resulting in the formation of unreliable links. To tackle this problem, we propose a novel multi-granular adaptive graph learning method designed to enhance the weight of causal features while reducing the weight of non-causal features across different granularities in constructing similarity graphs. Specifically, we first introduce a causally-guided hierarchical feature clustering method, wherein the distances between features are computed based on the differences in their causal contributions as represented in $\bm{W}$. The number of groups $M$ is then automatically determined by using the Calinski-Harabasz (CH) criterion~\cite{maulik2002performance}. Additionally, we adaptively learn granularity-specific similarity matrices at different granularities, ensuring that samples similar in high-dimensional space are also similar in low-dimensional space. Furthermore, the final similarity matrix is derived through the weighted fusion of granularity-specific similarity matrices, where the weights are determined by the causal contributions of features at each granularity level. This can be formulated as follows: 

\begin{equation}\label{3.4}
\begin{aligned}
\min_{\bm{S}^{(m)},\bm{G}}&~\frac{1}{2}\sum_{m=1}^{M}\sum_{i,j=1}^{n}\!\left[\|\bm{X}_{\cdot i}^{T}\bm{R}^{(m)}\bm{W}-\bm{X}_{\cdot j}^{T}\bm{R}^{(m)}\bm{W}\|_{2}^{2}\mathnormal{S}_{ij}^{(m)}\right.\\
&\left.+\gamma_{m}\|\bm{S}^{(m)}\|_{\mathrm{F}}^{2}\right]\!-\operatorname{Tr}(\bm{G}^{\top}\sum_{m=1}^{M}\nu_{m}\bm{S}^{(m)})+\xi\|\bm{G}\|_{\mathrm{F}}^{2}\\
\text { s.t. }&\mathnormal{S}_{ij}^{(m)}\geq 0, \bm{1}^{\top}\bm{S}_{\cdot i}^{(m)}=1,\mathnormal{G}_{ij}\geq 0,\bm{1}^{\top}\bm{G}_{\cdot i}=1,
\end{aligned}
\end{equation}
where $\bm{S}^{(m)}$ denotes the similarity matrix at the $m$-th granularity, $\bm{G}$ represents the fused similarity matrix, $\gamma_{m}$ and $\xi$ are two regularization parameters, and $\bm{R}^{(m)} \in \mathbb{R}^{d \times d}$ is a diagonal matrix with $R_{ii}^{(m)}=1$ if $i \in \Delta_{m}$, and 0 otherwise. Here, the set $\Delta_{m}$ denotes the index of features in the $m$-th granularity. Besides, $\nu_{m}$ is the weight of the $m$-th granularity, calculated as $\nu_{m}\!=\!\sum_{i\in\Delta_{m}}\!\|\bm{W}_{i\cdot}\|_{2}^{2}/\sum_{i=1}^{d}\|\bm{W}_{i\cdot}\|_{2}^{2}$. Eq.~(\ref{3.4}) shows that the more significant the contribution of causal features at the $m$-th granularity, the greater their weight in the final fused similarity graph.

By incorporating Eq. (\ref{3.4}) into Eq.~(\ref{3.3}), the final objective function of the proposed CAUSE-FS is formulated as:
\begin{equation}\label{3.5}
\begin{aligned}
\min_{\bm{\Theta}}&\!\sum_{i=1}^{n}\alpha\|\mu_{i}\bm{X}_{\cdot i}^{\top}\bm{W} \!-\! \bm{F}_{i \cdot} \|_{\mathrm{2}}^{2}+\operatorname{Tr}( \bm{F}^{\top}\bm{L}_{G}\bm{F})+\lambda\|\bm{W}\|_{2,1}\\
&+\!\frac{1}{2}\!\!\sum_{m=1}^{M}\!\sum_{i,j=1}^{n}\!\!\left[\!\|\mu_{i}\bm{X}_{\cdot i}^{T}\bm{R}^{(m)}\bm{W}\!-\!\mu_{j}\bm{X}_{\cdot j}^{T}\bm{R}^{(m)}\bm{W}\|_{2}^{2}\mathnormal{S}_{ij}^{(m)}\right.\\
&\left.+ \gamma_{m}\|\bm{S}^{(m)}\|_{\mathrm{F}}^{2}\right] -\operatorname{Tr}(\bm{G}^{\top}\sum_{m=1}^{M}\nu_{m}\bm{S}^{(m)})+\xi\|\bm{G}\|_{\mathrm{F}}^{2}\\
&+\beta \sum_{r=1}^{d}\| \frac{1}{|\Lambda_{t}|}\!\sum_{i\in\Lambda_{t}}\!\phi(\mu_{i}Z_{\cdot i}^{r}) - \frac{1}{|\Lambda_{c}|}\!\sum_{j\in\Lambda_{c}}\!\phi(\mu_{j}Z_{\cdot j}^{r})\|_{\mathcal{H}_{\kappa}}^{2}\\
\text { s.t. } &\!\bm{F}^{\top}\!\bm{F}\!=\!\bm{I},\mathnormal{S}_{ij}^{(m)}\!\geq\!0, \bm{1}^{\top}\!\bm{S}_{\cdot i}^{(m)}\!=\!1,\mathnormal{G}_{ij}\!\geq\!0, \bm{1}^{\top}\!\bm{G}_{\cdot i}\!=\!1,\\
&\mu_{i} \!\geq\! 0,\bm{1}^{\top}\bm{\mu}=1,
\end{aligned}
\end{equation}
where $\bm{\Theta}=\{\bm{W},\bm{F},\bm{G},\{\bm{S}^{(m)}\}_{m=1}^{M},\bm{\mu}\}$.

Eq.~(\ref{3.5}) shows two important advantages of the proposed CAUSE-FS method. On the one hand, unlike existing unsupervised feature selection approaches that focus solely on identifying discriminative features while neglecting causal aspects, CAUSE-FS is designed to identify causally informative features by jointly optimizing feature selection and confounder balancing. On the other hand, by considering the differing impacts of causal and non-causal features on the construction of the similarity graph, CAUSE-FS adaptively learns a more reliable similarity-induced graph matrix. This not only more effectively preserves the local structure of the data but also improves feature selection performance.

\section{Optimization}\label{sec:optimization}
Since the objective function in Eq.~(\ref{3.5}) is not jointly convex with respect to all variables, we develop an iterative algorithm to solve this optimization problem, in which each variable is alternately optimized while keeping the others fixed.

\subsection{Updating $\bm{W}$ by Fixing Others}
When other variables are fixed, $\bm{W}$ can be updated by solving the following optimization problem:
\begin{equation}\label{newW.1}
\begin{aligned}
\min_{\bm{W}}&\frac{1}{2}\sum_{m=1}^{M}\sum_{i,j=1}^{n}\|\mu_{i}\bm{X}_{\cdot i}^{T}\bm{R}^{(m)}\bm{W}-\mu_{j}\bm{X}_{\cdot j}^{T}\bm{R}^{(m)}\bm{W}\|_{2}^{2}\mathnormal{S}_{ij}^{(m)}\\
&+\sum_{i=1}^{n}\alpha\|\mu_{i}\bm{X}_{\cdot i}^{\top}\bm{W} - \bm{F}_{i \cdot} \|_{\mathrm{2}}^{2}+\lambda\|\bm{W}\|_{2,1}.\\
\end{aligned}
\end{equation}

According to the theorem proposed in~\cite{li2018generalized}, Eq.~(\ref{newW.1}) can be transformed into the following equivalent form:
\begin{equation}\label{W.2}
\begin{aligned}
\min_{\bm{W}}&\sum_{m=1}^{M}\operatorname{Tr}(\bm{W}^{\top}\bm{R}^{(m)}\tilde{\bm{X}}\bm{L}_{\mathnormal{S}}^{(m)}\tilde{\bm{X}}^{\top}\bm{R}^{(m)}\bm{W})\\
&+\alpha\|\tilde{\bm{X}}^{\top}\bm{W} - \bm{F}\|_{\mathrm{F}}^{2}+\lambda\operatorname{Tr}(\bm{W}^{\top}\bm{D}\bm{W}),\\
\end{aligned}
\end{equation}
where $\tilde{\bm{X}}=[\mu_{1}\bm{X}_{\cdot 1},\mu_{2}\bm{X}_{\cdot 2},\dots,\mu_{n}\bm{X}_{\cdot n} ]\in \mathbb{R}^{d \times n}$, $\bm{D}$ is a diagonal matrix with ${D}_{ii} \!=\! 1 / 2\sqrt{\|\bm{W}_{i \cdot}\|_{2}^{2}+\epsilon}$, and $\epsilon$ is a small constant to prevent the denominator from vanishing. 

By taking the derivative of Eq. (\ref{W.2}) w.r.t. $\bm{W}$ and setting it to 0, we obtain the optimal solution for $\bm{W}$ as follows:
\begin{equation}\label{W.3}
\begin{aligned}
\bm{W} \!\!=\!\! ( \alpha\tilde{\bm{X}}\tilde{\bm{X}}^{\top}\!\!+\!\lambda\bm{D}\!+\!\!\!\sum_{m=1}^{M}\!\!\bm{R}^{(m)}\!\tilde{\bm{X}}\bm{L}_{\mathnormal{S}}^{(m)}\tilde{\bm{X}}^{\top}\bm{R}^{(m)} )^{-1}\alpha\tilde{\bm{X}}\bm{F}.    
\end{aligned}
\end{equation}

\subsection{Updating $\bm{S}^{(m)}$ by Fixing Others}
While fixing other variables, the optimization problem w.r.t. $\bm{S}^{(m)}$ can be rewritten as follows:
\begin{equation}\label{newS.1}
\begin{aligned}
\min_{\bm{S}^{(m)}}&\frac{1}{2}\sum_{i,j=1}^{n}\|\mu_{i}\bm{X}_{\cdot i}^{T}\bm{R}^{(m)}\bm{W}-\mu_{j}\bm{X}_{\cdot j}^{T}\bm{R}^{(m)}\bm{W}\|_{2}^{2}\mathnormal{S}_{ij}^{(m)}\\
&+ \gamma_{m}\|\bm{S}^{(m)}\|_{\mathrm{F}}^{2} -\nu_{m}\operatorname{Tr}(\bm{G}^{\top}\bm{S}^{(m)})\\
\text { s.t. } &\mathnormal{S}_{ij}^{(m)}\geq0, \bm{1}^{\top}\bm{S}_{\cdot i}^{(m)}=1.\\
\end{aligned}
\end{equation}

Since Eq.~(\ref{newS.1}) is independent across columns, $\bm{S}^{(m)}$ can be optimized by separately solving each $\bm{S}_{\cdot j}^{(m)}$ as follows:
\begin{equation}\label{S.1}
\begin{aligned}
\min_{\bm{S}_{\cdot j}^{(m)}}&\frac{1}{2}\sum_{j=1}^{n}\|\mu_{i}\bm{X}_{\cdot i}^{T}\bm{R}^{(m)}\bm{W}-\mu_{j}\bm{X}_{\cdot j}^{T}\bm{R}^{(m)}\bm{W}\|_{2}^{2}\mathnormal{S}_{ij}^{(m)}\\
&-\nu_{m}\sum_{j=1}^{n}\mathnormal{G}_{ij}\mathnormal{S}_{ij}^{(m)}+ \gamma_{m}\|\bm{S}_{\cdot j}^{(m)}\|_{2}^{2}\\
\text { s.t. } &\mathnormal{S}_{ij}^{(m)}\!\geq\!0, \bm{1}^{\top}\!\bm{S}_{\cdot j}^{(m)}\!=\!1.
\end{aligned}
\end{equation}

By defining a vector $\bm{\tau}_{j}^{(m)}\in \mathbb{R}^{n \times 1}$ with the $i$-th entry as $\mathnormal{\tau}^{(m)}_{ij}=\frac{1}{2}\|\mu_{i}\bm{X}_{\cdot i}^{T}\bm{R}^{(m)}\bm{W}-\mu_{j}\bm{X}_{\cdot j}^{T}\bm{R}^{(m)}\bm{W}\|_{2}^{2}-\nu_{m}\mathnormal{G}_{ij}$, problem (\ref{S.1}) can be rewritten as:
\begin{equation}\label{S.2}
\begin{aligned}
&\min_{\bm{S}_{\cdot j}^{(m)}}\frac{1}{2}\| \bm{S}_{\cdot j}^{(m)} + \bm{\tau}_{j}^{(m)}/2\gamma_{m} \|_{2}^{2}\\
&\text { s.t. }\mathnormal{S}_{ij}^{(m)}\!\geq\!0, \bm{1}^{\top}\!\bm{S}_{\cdot j}^{(m)}\!=\!1.
\end{aligned}
\end{equation}

The Lagrangian function of the problem~(\ref{S.2}) can be formulated as follows:
\begin{equation}\label{newS.2}
\mathcal{L}(\bm{S}_{\cdot j}^{(m)}\!\!,\bm{\Psi},\eta)\!=\! \frac{1}{2}\| \bm{S}_{\cdot j}^{(m)} \!+ \frac{\bm{\tau}_{j}^{(m)}}{2\gamma_{m}} \|_{2}^{2}\!-\!\bm{\Psi}^{\top}\!\bm{S}_{\cdot j}^{(m)}\!\!-\!\eta(\bm{1}^{\top}\!\bm{S}_{\cdot j}^{(m)}\!-\!1),
\end{equation}
where $\bm{\Psi}$ and $\eta$ represent the Lagrange coefficient vector and scalar, respectively.

By taking the derivative of Eq. (\ref{newS.2}) w.r.t $\bm{S}_{\cdot j}^{(m)}$ and setting it to zero, we get:
\begin{equation}\label{newS.3}
\bm{S}_{\cdot j}^{(m)} + {\bm{\tau}_{j}^{(m)}}/{2\gamma_{m}} - \bm{\Psi}^{\top}-\eta\bm{1}^{\top} = \bm{0}.
\end{equation}

Based on the Karush-Kuhn-Tucker (KKT) complementary conditions~\cite{gordon2012karush}, i.e., ${S}_{i j}^{(m)}{\Psi}_{i}=0$, the solution for ${S}_{i j}^{(m)}$ can be derived as:
\begin{equation}\label{newS.4}
\hat{\mathnormal{S}}_{ij}^{(m)}=\max(\eta - {\tau_{ij}^{(m)}}/{2\gamma_{m}},0).
\end{equation}

To ensure that $\bm{S}^{(m)}$ can characterize the local structure of data, we constrain $\bm{S}_{\cdot j}^{(m)}$ to contain only $k$ non-zero elements, where $k$ represents the number of nearest neighbors. To this end, we first arrange $\tau_{1j}^{(m)} \dots \tau_{nj}^{(m)} $ in ascending order, then set $\hat{\mathnormal{S}}_{kj}^{(m)}>0$ and $\hat{\mathnormal{S}}_{k+1,j}^{(m)}=0$ to guarantee that $\bm{S}_{\cdot j}^{(m)}$ only have $k$ nonzero entries. Hence, we have $\eta - {\tau_{kj}^{(m)}}/{2\gamma_{m}}>0$ and $\eta - {\tau_{k+1,j}^{(m)}}/{2\gamma_{m}} \leq 0$. Combined with the constraint $\bm{1}^{T} \bm{S}_{\cdot j}^{(m)}=1$, we can get $\eta = \frac{1}{k}+\frac{1}{2k\gamma_{m}}\sum_{p=1}^{k}\tau_{pj}^{(m)}$. Thus, let $\gamma_{m}=(k \tau_{k+1,j}^{(m)}-\sum_{p=1}^{k} \tau_{pj}^{(m)})/2$, we can obtain the optimal solution of $\mathnormal{S}_{ij}^{(m)}$ as follows:
\begin{equation}\label{S.5}
\mathnormal{S}_{i j}^{(m)}=\left\{\begin{array}{cc}
\frac{\tau_{k+1,j}^{(m)}-\tau_{i j}^{(m)}}{k \tau_{k+1,j}^{(m)}-\sum_{p=1}^{k} \tau_{pj}^{(m)}} & j \leq k; \\
0 & j>k.
\end{array}\right.
\end{equation}

\subsection{Updating $\bm{G}$ by Fixing Others} 
With other variables fixed, the optimization problem w.r.t. $\bm{G}$ becomes:
\begin{equation}\label{newG.1}
\begin{aligned}
\min_{\bm{G}}&\operatorname{Tr}( \bm{F}^{\top}\bm{L}_{G}\bm{F}) -\operatorname{Tr}(\bm{G}^{\top}\sum_{m=1}^{M}\nu_{m}\bm{S}^{(m)})+\xi\|\bm{G}\|_{\mathrm{F}}^{2}\\
\text { s.t. } &\mathnormal{G}_{ij}\!\geq\!0, \bm{1}^{\top}\!\bm{G}_{\cdot i}\!=\!1.\\
\end{aligned}
\end{equation}

Based on the matrix trace property, we can transform problem (\ref{newG.1}) into:
\begin{equation}\label{G.2}
\begin{aligned}
\min_{\bm{G}}&\sum_{i,j=1}^{n}\frac{1}{2}\| \bm{F}_{i \cdot}-\bm{F}_{j \cdot} \|_{2}^{2}\mathnormal{G}_{ij}-\mathnormal{G}_{ij}\bar{{S}}_{ij}+\xi\mathnormal{G}_{ij}^{2}\\
\text { s.t. } &\mathnormal{G}_{ij}\geq 0, \bm{1}^{\top}\bm{G}_{\cdot i}=1,
\end{aligned}
\end{equation}
where $\bar{\bm{S}}=\sum_{m=1}^{M}\nu_{m}\bm{S}^{(m)}$.

Similar to the process of solving Eq.~(\ref{newS.1}), the optimal solution for $\bm{G}$  can be obtained by
\begin{equation}\label{G.3}
\mathnormal{G}_{i j}=\left\{\begin{array}{cc}
\frac{\delta_{k+1,j}-\delta_{i j}}{k \delta_{k+1,j}-\sum_{p=1}^{k} \delta_{pj}} & j \leq k; \\
0 & j>k,
\end{array}\right.
\end{equation}
where $\mathnormal{\delta}_{ij}=\frac{1}{2}\| \bm{F}_{i \cdot}-\bm{F}_{j \cdot} \|_{2}^{2}-\bar{S}_{ij}$, and $\xi$ is determined by $\xi=(k \delta_{k+1,j}-\sum_{p=1}^{k} \delta_{pj})/2$.

\subsection{Updating $\bm{F}$ by Fixing Others}
When fixing other variables, problem (\ref{3.5}) can be reformulated as:
\begin{equation}\label{F.1}
\begin{aligned}
\min_{\bm{F}}&\sum_{i=1}^{n}\alpha\|\mu_{i}\bm{X}_{\cdot i}^{\top}\bm{W} - \bm{F}_{i \cdot} \|_{\mathrm{2}}^{2}+\operatorname{Tr}( \bm{F}^{\top}\bm{L}_{G}\bm{F})\\
\text { s.t. } &\bm{F}^{\top}\bm{F}=\bm{I}.
\end{aligned}
\end{equation}

By removing the terms irrelevant to $\bm{F}$, problem (\ref{F.1}) can be rewritten as:
\begin{equation}\label{F.2}
\begin{aligned}
\min_{\bm{F}}&\operatorname{Tr}(\bm{F}^{\top}(\alpha\bm{I}+\bm{L}_{G})\bm{F}-2\alpha\bm{F}^{\top}\tilde{\bm{X}}\bm{W})\\
\text { s.t. } &\bm{F}^{\top}\bm{F}=\bm{I}.
\end{aligned}
\end{equation}

Problem (\ref{F.2}) can be efficiently solved by the generalized power iteration (GPI) method~\cite{nie2017generalized}.

\subsection{Updating $\bm{\mu}$ by Fixing Others}
After fixing other variables, the objective function w.r.t $\bm{\mu}$ is reduced to:
\begin{equation}\label{newu.1}
\begin{aligned}
\min_{\bm{\mu}}&\frac{1}{2}\sum_{m=1}^{M}\sum_{i,j=1}^{n}\|\mu_{i}\bm{X}_{\cdot i}^{T}\bm{R}^{(m)}\bm{W}-\mu_{j}\bm{X}_{\cdot j}^{T}\bm{R}^{(m)}\bm{W}\|_{2}^{2}\mathnormal{S}_{ij}^{(m)}\\
&+\beta \sum_{r=1}^{d}\| \frac{1}{|\Lambda_{t}|}\!\sum_{i\in\Lambda_{t}}\!\phi(\mu_{i}Z_{\cdot i}^{r}) - \frac{1}{|\Lambda_{c}|}\!\sum_{j\in\Lambda_{c}}\!\phi(\mu_{j}Z_{\cdot j}^{r})\|_{\mathcal{H}_{\kappa}}^{2}\\
&+\sum_{i=1}^{n}\alpha\|\mu_{i}\bm{X}_{\cdot i}^{\top}\bm{W} \!-\! \bm{F}_{i \cdot} \|_{\mathrm{2}}^{2}\\
\text { s.t. } &\mu_{i} \geq 0,\bm{1}^{\top}\bm{\mu}=1,
\end{aligned}
\end{equation}

Problem (\ref{newu.1}) can be reformulated as follows by utilizing the empirical estimation of maximum mean discrepancy from~\cite{kumagai2019unsupervised}:
\begin{equation}\label{u.2}
\begin{aligned}
&\min_{\bm{\mu}}\sum_{i,j=1}^{n}\sum_{m=1}^{M}\frac{1}{2}\mu_{i}^{2}\mathnormal{P}_{ii}^{(m)}(\mathnormal{S}_{ij}^{(m)}+\mathnormal{S}_{ji}^{(m)})-\mu_{i}\mu_{j}\mathnormal{P}_{ij}^{(m)}\mathnormal{S}_{ij}^{(m)}\\
&+\frac{\beta}{n^2}\!\sum_{i,j=1}^{n}\sum_{r=1}^{d}\mu_{i}\mu_{j}(\mathnormal{E}_{ri}\mathnormal{E}_{rj}\!-\!2\mathnormal{E}_{ri}\mathnormal{C}_{rj}\!+\!\mathnormal{C}_{ri}\mathnormal{C}_{rj})\bm{Z}_{\cdot i}^{r \top}\!\bm{Z}_{\cdot j}^{r}\\
&+\sum_{i=1}^{n}\alpha\mu_{i}^{2}\mathnormal{B}_{ii}-2\alpha\mu_{i}\bm{X}_{\cdot i}^{\top}\bm{W}\bm{F}_{i\cdot}^{\top}+\alpha\bm{F}_{i\cdot}\bm{F}_{i\cdot}^{\top}\\
&\text { s.t. } \mu_{i} \geq 0, \textbf{\textit{1}}^{\top}\bm{\mu}=1,\\
\end{aligned}
\end{equation}
where $\mathnormal{E}_{rj}$ is an indicator variable whose value is 1, indicating that the $j$-th sample is assigned to the treatment group when the $r$-th feature is a treatment feature; otherwise, this sample is assigned to the control group. Besides, $\mathnormal{C}_{rj} = 1-\mathnormal{E}_{rj}$, $\mathnormal{P}^{(m)}_{ij}=\bm{X}_{\cdot i}^{\top}\bm{R}^{(m)}\bm{W}\bm{W}^{\top}\bm{R}^{(m)}\bm{X}_{\cdot j}$, and $\mathnormal{B}_{ij}=\bm{X}_{\cdot i}^{\top}\bm{W}\bm{W}^{\top}\bm{X}_{\cdot j}$. 

We can transform problem (\ref{u.2}) into the following equivalent problem:
\begin{equation}\label{u.3}
\begin{aligned}
&\min_{\bm{\mu}}\sum_{i=1}^{n}(\tilde{b}_{i}+\mathnormal{\Gamma}_{ii})\mu_{i}^{2}+\sum_{i=1}^{n}\sum_{j\neq i}^{n}\mathnormal{\Gamma}_{ij}\mu_{i}\mu_{j}+\sum_{i=1}^{n}\tilde{a}_{i}\mu_{i} \\
&\text { s.t. } \mu_{i} \geq 0, \bm{1}^{\top}\bm{\mu}=1,\\
\end{aligned}
\end{equation}
where $\tilde{b}_{i}=\alpha\mathnormal{B}_{ii}+\frac{1}{2}\sum_{j=1}^{n}\sum_{m=1}^{M}\mathnormal{P}_{ii}^{(m)}(\mathnormal{S}_{ij}^{(m)}+\mathnormal{S}_{ji}^{(m)})$, $\mathnormal{\Gamma}_{ij}=\frac{\beta}{n^2}\sum_{r=1}^{d}\mathnormal{Q}_{ij}^{r}\bm{Z}_{\cdot i}^{r \top}\bm{Z}_{\cdot j}^{r}-\sum_{m=1}^{M}\mathnormal{P}_{ij}^{(m)}\mathnormal{S}_{ij}^{(m)}$, $\mathnormal{Q}_{ij}^{r}=\mathnormal{E}_{ri}\mathnormal{E}_{rj}-2\mathnormal{E}_{ri}\mathnormal{C}_{rj}+\mathnormal{C}_{ri}\mathnormal{C}_{rj}$, and $\tilde{a}_{i}=-2\alpha\bm{X}_{\cdot i}^{\top}\bm{W}\bm{F}_{i \cdot}^{\top}$. 

Problem (\ref{u.3}) is a standard quadratic objective function with linear constraints, which can be efficiently solved by the interior-point method~\cite{potra2000interior}.

\begin{algorithm}[t]
	\caption{Iterative Algorithm of CAUSE-FS}
	\textbf{Input:} The data matrix $\bm{X}\!\in\! \mathbb{R}^{d \times n}$, parameters $\alpha$, $\beta$ and $\lambda$. 
	
	1: \parbox[t]{\textwidth}{\textbf{Initialize:} $\bm{Z}^{i}$, $\Delta_{m}$, $\bm{S}^{(m)}$, $\bm{G}$, $\bm{F}$, $\bm{\mu}$, $\bm{\nu}$, and $\bm{D} = \bm{I}$.} 
	
	2: \textbf{While} {not convergent} \textbf{do}
	
	3: \quad Update $\bm{W}$ via Eq. (\ref{W.3});
	
	4: \quad Update $\mathnormal{D}_{ii} = 1 / 2\sqrt{\|\bm{W}_{i \cdot}\|_{2}^{2}+\epsilon}$;
	
	5: \quad Update $\{\bm{S}^{(m)}\}_{m=1}^{M}$ via Eq. (\ref{S.5});
	
	6: \quad Update $\bm{G}$ via Eq. (\ref{G.3});
	
	7: \quad Update $\bm{F}$ by solving problem (\ref{F.2});
	
	8: \quad Update $\bm{\mu}$ by solving problem (\ref{u.3});
	
	9: \quad \parbox[c]{\textwidth}{Update $\{\Delta_{m}\}_{m=1}^{M}$ using causally-guided hierarchical \\  feature clustering;}
	
	10: \hspace{0.65mm} {Update $\{\nu_{m}\!=\!\sum_{i\in\Delta_{m}}\!\|\bm{W}_{i\cdot}\|_{2}^{2}/\sum_{i=1}^{d}\|\bm{W}_{i\cdot}\|_{2}^{2}\}_{m=1}^{M}$.}
	
	11: \hspace{1.3mm} \parbox[c]{\textwidth}{Update $\bm{Z}^{i}$ by removing irrelevant and adjustment \\variables based on $\bm{W}$.}
	
	12: \textbf{End}
	
	\textbf{Output:} {Sorting the $\ell_{2}$-norm of rows of $\bm{W}$ in a descending order and selecting the top $\rho$ features.}
\end{algorithm}

Algorithm 1 summarizes the detailed optimization procedure of CAUSE-FS. Within this algorithm, $\{\Delta_{m}\}_{m=1}^{M}$ is initialized via  hierarchical feature clustering based on the Pearson correlation coefficient~\cite{Benesty2009}. $M$ is determined automatically using the CH criterion and $\nu_{m}$ is set as $1/M$ for all feature granularities. $\{\bm{S}^{(m)}\}_{m=1}^{M}$ and $\bm{G}$ are initialized by constructing $k$-nearest neighbor graph according to~\cite{wang2019gmc}. $\bm{F}$ is initialized through spectral clustering~\cite{bach2003learning}. $\mu_{i}$ is set as $1/n$ for all samples.

\section{Convergence and Complexity Analysis}\label{sec:Discussions}
In this section, we provide a theoretical analysis of the convergence and computational complexity of the proposed CAUSE-FS algorithm.

\subsection{Convergence Analysis}\label{Convergence Analysis}
Since the objective function in Eq. (\ref{3.5}) is not simultaneously convex with respect to the five variables $\bm{W}$, $\bm{S}^{(m)}$, $\bm{G}$, $\bm{F}$, and $\bm{\mu}$, we decompose it into five sub-objective functions: Eqs. (\ref{newW.1}), (\ref{newS.1}), (\ref{newG.1}), (\ref{F.1}), and (\ref{newu.1}). To prove the convergence of Algorithm 1, we demonstrate the monotonic convergence of each sub-objective function.  Before delving into the details, we first introduce the following lemma proposed by~\cite{hou2013joint}.

\begin{lemma}\label{lemma1}
For any two nonzero vectors $\bm{a}$ and $\bm{b}$, the following inequality is satisfied:
\begin{equation}
\begin{aligned}
\|\bm{a}\|_{2}-\frac{\|\bm{a}\|_{2}^{2}}{2\|\bm{b}\|_{2}} \leq \|\bm{b}\|_{2}-\frac{\|\bm{b}\|_{2}^{2}}{2\|\bm{b}\|_{2}}
\end{aligned}
\end{equation}
\end{lemma}

We begin by presenting the following theorem to demonstrate the decrease of the objective function in Eq. (\ref{newW.1}) when updating $\bm{W}$ while keeping other variables fixed.
\begin{theorem}\label{theorem1}
Through the updating rules in Algorithm 1,  the objective function of CAUSE-FS in Eq. (\ref{newW.1}) is non-increasing.
\end{theorem}

\begin{proof}
	Since $\bm{W}_{t+1}$ is derived as the optimal solution to the problem (\ref{W.2}), it holds that
	\begin{equation}\label{Conver.1}
		\begin{aligned}
			&\operatorname{Tr}(\bm{W}_{\!t+1}^{\top}\!\bm{V}\bm{W}_{\!t+1}\!)\!+\!\alpha\|\tilde{\bm{X}}^{\top}\!\bm{W}_{\!t+1} \!\!-\! \bm{F}\|_{\mathrm{F}}^{2}\!+\!\lambda\operatorname{Tr}(\bm{W}_{\!t+1}^{\top}\bm{D}_{t}\bm{W}_{\!t+1})\\
			&\leq \operatorname{Tr}(\bm{W}_{t}^{\top}\bm{V}\bm{W}_{t})+\alpha\|\tilde{\bm{X}}^{\top}\bm{W}_{t} - \bm{F}\|_{\mathrm{F}}^{2}+\lambda\operatorname{Tr}(\bm{W}_{t}^{\top}\bm{D}_{t}\bm{W}_{t}),
		\end{aligned}
	\end{equation}
	where $\bm{V}=\sum_{m=1}^{M}\bm{R}^{(m)}\tilde{\bm{X}}\bm{L}_{\mathnormal{S}}^{(m)}\tilde{\bm{X}}^{\top}\bm{R}^{(m)}$.
	
	According to Lemma~\ref{lemma1}, we can get
	\begin{equation}\label{Conver.2}
		\begin{aligned}
			& \|{\bm{w}}_{t+1}^{i}\|_2- \frac{\|{\bm{w}}_{t+1}^{i}\|_{2}^{2}}{2\|{\bm{w}}_{t}^{i}\|_2} \leq \|\bm{w}_{t}^{i}\|_2- \frac{\|\bm{w}_{t}^{i}\|_{2}^{2}}{2\|\bm{w}_{t}^{i}\|_2}\\
			\Rightarrow &\sum_{i=1}^{d}(\|{\bm{w}}_{t+1}^{i}\|_2-\frac{\|{\bm{w}}_{t+1}^{i}\|_{2}^{2}}{2\|{\bm{w}}_{t}^{i}\|_2}) \leq \sum_{i=1}^{d}(\|\bm{w}_{t}^{i}\|_2-\frac{\|\bm{w}_{t}^{i}\|_{2}^{2}}{2\|\bm{w}_{t}^{i}\|_2}), \\
			\Rightarrow&\|\!{\bm{W}}_{t+1}\!\|_{2,1}\!\!-\!\!\operatorname{Tr}({\bm{W}}_{t+1}^{\top}\!\bm{D}_{t}{\bm{W}}_{t+1})\!\leq\!\|\!\bm{W}_{t}\!\|_{2,1}\!\!-\!\!\operatorname{Tr}(\bm{W}_{t}^{\top}\!\bm{D}_{t}\bm{W}_{t}),
		\end{aligned}
	\end{equation}
	where $\bm{w}_{t+1}^{i}$ denotes the $i$-th row of $\bm{W}_{t+1}$.
	
	By combining Eqs.~(\ref{Conver.1}) and (\ref{Conver.2}), we have:
	\begin{equation}\label{Conver.3}
		\begin{aligned}
			&\operatorname{Tr}(\bm{W}_{\!t+1}^{\top}\!\bm{V}\bm{W}_{\!t+1}\!)\!+\!\alpha\|\tilde{\bm{X}}^{\top}\!\bm{W}_{\!t+1} \!\!-\! \bm{F}\|_{\mathrm{F}}^{2}\!+\!\lambda\|{\bm{W}}_{t+1}\|_{2,1}\\
			&\leq \operatorname{Tr}(\bm{W}_{t}^{\top}\bm{V}\bm{W}_{t})+\alpha\|\tilde{\bm{X}}^{\top}\bm{W}_{t} - \bm{F}\|_{\mathrm{F}}^{2}+\lambda\|{\bm{W}}_{t}\|_{2,1}.
		\end{aligned}
	\end{equation}
	Thus, the objective function value of Eq.~(\ref{newW.1}) decreases monotonically with each iteration of Algorithm 1.
\end{proof}

The convergences of updating $\bm{S}^{(m)}$ and $\bm{G}$ can be guaranteed by their closed solutions in Eqs.~(\ref{S.5}) and (\ref{G.3}), respectively. Additionally, the convergence of updating $\bm{F}$ follows from the GPI method\cite{nie2017generalized}. Furthermore, updating $\bm{\mu}$ is proven to converge when the interior point method is used to solve the linear quadratic objective function in Eq.~(\ref{u.3})~\cite{boyd2004convex}. Consequently, the objective function in Eq.~(\ref{3.5}) converges based on the update rules in (\ref{W.3}), (\ref{S.5}), (\ref{G.3}), (\ref{F.2}), and (\ref{u.3}). Finally, the convergence behavior of Algorithm 1 will be experimentally validated in the experiment section.

\subsection{Time Complexity Analysis} 
In each iteration of Algorithm 1, the computational complexity for updating $\bm{W}$ is $\mathcal{O}(d^2max(d,n)+dn^2)$. Both $\bm{S}^{(m)}$ and $\bm{G}$ updates incur a computational cost of $\mathcal{O}(n^2logn)$. The update of $\bm{F}$ via the GPI algorithm requires $\mathcal{O}(n^2h)$. The quadratic programming solution for updating $\bm{\mu}$ has a complexity of $\mathcal{O}({n}^{3})$. Hence, the overall computational cost per iteration in Algorithm 1 is $\mathcal{O}(d^2\max(d,n)+n^2\max(d,n))$.

\section{Experiments}\label{Experiments}
\subsection{Experimental Settings}
\subsubsection{Datasets}
The experiments are conducted on six real-world datasets, including one biological dataset (Lung\footnote{https://jundongl.github.io/scikit-feature/datasets.html}), two face image datasets (Jaffe\footnote{https://hyper.ai/datasets/17880} and Umist\footnote{https://www.openml.org/}), one object image dataset (COIL20\footnotemark[3]), two handwritten digit datasets (Gisette\footnotemark[3] and USPS\footnotemark[3]). The statistical information of these datasets is summarized in Table~\ref{Data}. Details of these datasets are provided as follows.

\textit{Lung:} It consists of 203 samples classified into five distinct lung cancer subtypes: 139 adenocarcinoma samples, 17 normal lung samples, 6 small cell lung cancer samples, 21 squamous cell carcinoma samples, and 20 carcinoid lung cancer samples. Each sample is described by 3312 features.

\textit{Jaffe:} It is a facial expression recognition dataset containing 213 images from 10 individuals, each depicting one of seven expressions: sadness, happiness, anger, disgust, surprise, fear, or neutrality. Each image is represented as a 676-dimensional visual feature vector.

\textit{Umist:} It comprises 575 grayscale images of 20 individuals, with each individual's images capturing a variety of poses, from profile to frontal views. Each image is encoded as a 644-dimensional visual descriptor.

\textit{COIL20:} It is made up of 1440 grayscale images of 20 objects, including toys, household items, and vehicles. Each object is depicted in 72 grayscale images, all with a resolution of 32$\times$32 pixels.

\textit{Gisette:} It is a handwritten digit recognition dataset comprising 7000 images of the digits four and nine. Each image is represented by a 5000-dimensional feature vector, which includes pixel values from the original image as well as higher-order features derived from the products of these pixel values.

\textit{USPS:} It comprises 9,298 grayscale images of handwritten digits, representing the numbers 0 through 9. Each image has a resolution of 16$\times$16 pixels and is flattened into a 256-dimensional feature vector, where each dimension corresponds to the grayscale value of a pixel.

\begin{table}[t]
	\centering
	\normalsize
	\caption{Statistics of different datasets}\label{Data}
	\setlength{\tabcolsep}{4mm}
	\begin{tabular}{@{\extracolsep{\fill}}lccc}
		\toprule
		Datasets  & \# Samples & \# Features  & \# Classes\\
		% \hline
		\midrule
		Lung      & 203 &3312 & 5 \\
		
		Jaffe    & 213  &676 & 10 \\
		
		Umist      & 575 &644  & 20 \\
		
		COIL20  & 1440 &1024  & 20 \\
		
		Gisette      & 7000 & 5000  & 2 \\
		
		USPS    & 9298 & 256  & 10 \\
		\toprule
	\end{tabular}
\end{table}

\subsubsection{Compared Methods}
We compare the proposed CAUSE-FS  with several SOTA approaches, as well as a baseline method (AllFea) that utilizes all original features. A brief introduction to these methods is provided below:

\begin{itemize}
\item \textbf{AllFea} utilizes all original features for comparison.
\item \textbf{FDNFS}~\cite{FDNFS} leverages the manifold structure of the original data and feature correlations to guide the feature selection process.
\item \textbf{HSL}~\cite{HSL}  adaptively learns an optimal graph Laplacian by utilizing high-order similarity information.
\item \textbf{OEDFS}~\cite{OEDFS} integrates self-representation learning and non-negative matrix factorization to identify discriminative features.
\item \textbf{FSDK}~\cite{FSDK}  incorporates feature selection into the framework of the discriminative K-means clustering.
\item \textbf{BLFSE}~\cite{BLFSE} combines feature-level and clustering-level ensemble learning to perform feature selection.
\item \textbf{VCSDFS}~\cite{VCSDFS} proposes a variance-covariance subspace distance combined with inner product regularization to facilitate feature selection.
\item \textbf{DUFS}~\cite{DUFS} integrates both local and global feature correlations into the feature selection process.
\item \textbf{CNAFS}~\cite{CNAFS} introduces a unified feature selection framework that combines self-expression with pseudo-label learning.
\end{itemize}

\begin{table*}[!htbp]
	\centering
	\normalsize
	% \small
	\caption{Means and standard deviation (\%) of ACC for different methods on six datasets with the number of selected features fixed at 20.}\label{ClusteringACC}
	\setlength{\tabcolsep}{3mm}
	\begin{tabular}{lcccccc}
		\toprule
		{Methods} & {Lung}&{Jaffe} &{Umist} &{COLI20}&{Gisette} &{USPS}\\
		% \hline
		\midrule
		{CAUSE-FS} & \textbf{81.43 $\pm$ 7.29 } & \textbf{85.08 $\pm$ 8.69} & \textbf{55.99 $\pm$ 3.34} & \textbf{59.01 $\pm$ 3.73} & \textbf{83.32 $\pm$ 0.54} & \textbf{66.82 $\pm$ 1.26}\\
		
		{AllFea}  &69.59 $\pm$ 6.75& 77.47 $\pm$ 7.61 & 41.50 $\pm$ 2.81 & 56.90 $\pm$ 4.16 &  68.43 $\pm$ 0.12 & 64.17 $\pm$ 2.98  \\
		
		{FDNFS}  &67.22 $\pm$ 5.91 & 66.61 $\pm$ 4.98 & 44.00 $\pm$ 2.99 &  49.67 $\pm$ 2.64 & 50.05 $\pm$ 0.01  & 49.78 $\pm$ 3.28\\
		
		{HSL}  &63.17 $\pm$ 5.32& 77.69 $\pm$ 6.04 & 49.79 $\pm$ 2.75 & 53.90 $\pm$ 3.29 & 71.18 $\pm$ 3.45 & 58.79 $\pm$ 3.92\\
		
		{OEDFS}  &57.90 $\pm$ 5.63& 64.24 $\pm$ 5.25& 42.13 $\pm$ 2.68 & 49.55 $\pm$ 2.40 & 56.08 $\pm$ 2.23 & 46.07 $\pm$ 1.68\\
		
		{FSDK}  &67.06 $\pm$ 4.21& 78.43 $\pm$ 4.52  & 46.31 $\pm$ 2.31  & 57.64 $\pm$ 1.97  & 74.06 $\pm$ 0.12   & 62.65 $\pm$ 0.56 \\
		
		{BLFSE}  &51.57 $\pm$ 5.71 & 58.31 $\pm$ 3.27& 41.06 $\pm$ 2.57 & 53.21 $\pm$ 2.36 & 60.50 $\pm$ 0.01 &  52.79 $\pm$ 1.29\\
		
		{VCSDFS}  &55.96 $\pm$ 6.72 & 72.64 $\pm$ 5.88 & 42.32 $\pm$ 2.96 & 56.76 $\pm$ 3.35  & 58.61 $\pm$ 0.04 & 56.81 $\pm$ 4.22  \\
		
		{DUFS}  &72.69 $\pm$ 8.51 & 72.10 $\pm$ 3.27 & 47.54 $\pm$ 4.48  & 50.19 $\pm$ 2.71  &  54.21 $\pm$ 0.06  &  62.77 $\pm$ 0.28 \\
		
		{CNAFS}  &64.32 $\pm$ 7.55 & 76.68 $\pm$ 6.81& 48.01 $\pm$2.92  & 56.63 $\pm$ 3.66  & 57.89 $\pm$ 2.85  & 61.57 $\pm$ 3.73  \\
		\bottomrule
	\end{tabular}
\end{table*}

\begin{table*}[!htbp]
	\centering
	\normalsize
	% \small
	\caption{Means and standard deviation (\%) of NMI for different methods on six datasets with the number of selected features fixed at 20.}\label{ClusteringNMI}
	\setlength{\tabcolsep}{3mm}
	\begin{tabular}{lcccccc}
		\toprule
		{Methods} & {Lung}&{Jaffe} &{Umist} &{COLI20}&{Gisette} &{USPS}\\
		% \hline
		\midrule
		{CAUSE-FS}  & \textbf{63.40 $\pm$ 8.77}  & \textbf{89.26 $\pm$ 4.99}  & \textbf{72.10 $\pm$ 2.26} & \textbf{70.00 $\pm$ 1.77}  & \textbf{33.01 $\pm$ 0.91} & \textbf{60.31 $\pm$ 3.12}\\
		
		{AllFea}  & 50.99 $\pm$ 5.62  & 82.51 $\pm$ 4.74  & 62.96 $\pm$ 2.06  &  68.38 $\pm$ 1.96  &  11.67 $\pm$ 0.15  & 56.73 $\pm$ 0.84  \\
		
		{FDNFS}  & 29.93 $\pm$ 5.59  & 71.96 $\pm$ 2.80  & 55.08 $\pm$ 2.31  &  51.78 $\pm$ 1.56 & 4.01 $\pm$ 0.01    & 42.21 $\pm$ 1.03\\
		
		{HSL}  & 34.54 $\pm$ 3.09 & 80.19 $\pm$ 3.73 & 65.74 $\pm$ 2.06 &  64.48 $\pm$ 1.92  & 15.62 $\pm$ 3.10   &  51.55 $\pm$ 1.71\\
		
		{OEDFS}  & 27.87 $\pm$ 1.99 & 64.38 $\pm$ 3.80  &  57.01 $\pm$ 2.40  &  55.08 $\pm$ 1.43  &  1.40 $\pm$ 0.59   &  38.07 $\pm$ 0.68\\
		
		{FSDK}  & 46.86 $\pm$ 2.15 & 81.38 $\pm$ 5.67 & 63.89 $\pm$ 2.82  &  66.84 $\pm$ 2.84  & 23.01 $\pm$ 0.01   &  56.44 $\pm$ 0.62\\
		
		{BLFSE}  & 32.10 $\pm$ 3.42 & 62.85 $\pm$ 2.60 & 55.20 $\pm$ 1.64  &  62.00 $\pm$ 1.60  &  3.20 $\pm$ 0.01  &  43.12 $\pm$ 1.07\\
		
		{VCSDFS}  & 33.82 $\pm$ 4.04 & 77.94 $\pm$ 3.80 & 58.66 $\pm$ 2.05  & 67.67 $\pm$ 1.64   &  2.35 $\pm$ 0.02  &  46.49 $\pm$ 1.70\\
		
		{DUFS}  & 51.88 $\pm$ 6.84 & 75.41 $\pm$ 4.05 & 59.86 $\pm$ 3.96  &  60.73 $\pm$ 1.41  & 4.04 $\pm$ 0.03   & 52.79 $\pm$ 0.88\\
		
		{CNAFS}  & 46.77 $\pm$ 2.74 & 80.26 $\pm$ 4.23 & 60.95 $\pm$ 2.03 &  67.27 $\pm$ 1.80  &  4.35 $\pm$ 1.59  &  53.18 $\pm$ 1.32\\
		\bottomrule
	\end{tabular}
\end{table*}

\begin{figure*}[!htbp]
	\centering
	\includegraphics[width=\textwidth]{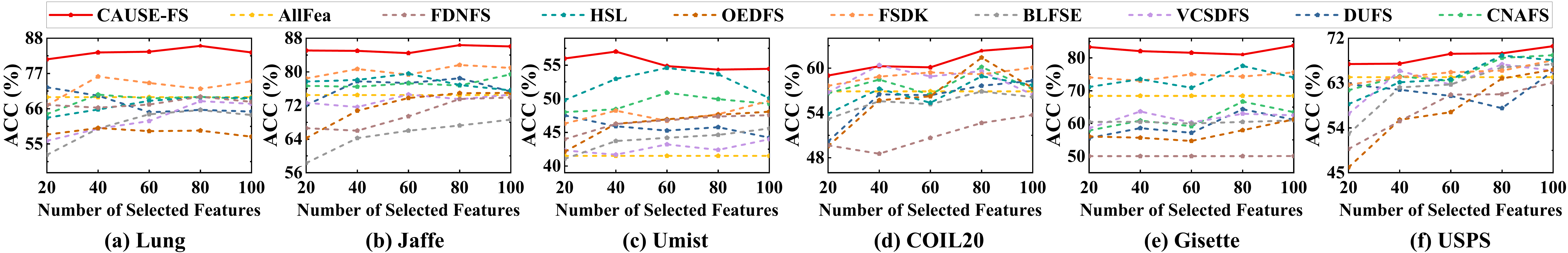} 
	\caption{ACC of different methods across various numbers of selected features on six datasets.}
	\label{ACC}
\end{figure*}

\begin{figure*}[!htbp]
	\centering
	\includegraphics[width=\textwidth]{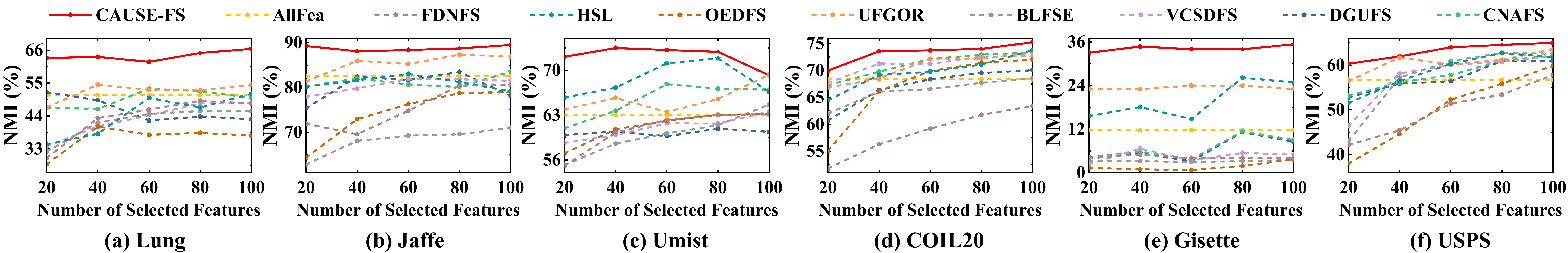} 
	\caption{NMI of different methods across various numbers of selected features on six datasets.}
	\label{NMI}
\end{figure*}

\subsubsection{Comparison Schemes}
To ensure a fair comparison,  we utilize a grid search strategy for parameter tuning across all compared methods and report the best performance. Meanwhile, the parameters $\alpha$ and $\lambda$ of our method are tuned over the range $\{ 10^{-3},10^{-2},10^{-1},1,10^{1},10^{2},10^{3} \}$, and $\beta$ is tuned within  $\{ 10^{5},10^{6},10^{7},10^{8},10^{9},10^{10} \}$. Since determining the optimal number of selected features is still challenging, we set it to range from 20 to 100, in increments of 20. In the evaluation phase, we follow a common way to assess UFS~\cite{li2019discriminative,chen2022fast}, by using two widely recognized clustering metrics: Clustering Accuracy (ACC)~\cite{ng2001spectral} and Normalized Mutual Information (NMI)~\cite{strehl2002cluster}, to evaluate the quality of the selected features. We conduct K-means clustering 50 times on the selected features and report the average results and standard deviations.

\subsection{Experimental Results and Analysis}
\subsubsection{Performance Comparisons}
Table~\ref{ClusteringACC} and Table~\ref{ClusteringNMI} summarize the performance of CAUSE-FS and other comparison methods in terms of ACC and NMI, respectively, with the best performance highlighted in boldface. As observed, CAUSE-FS consistently achieves the highest performance across all datasets and metrics when compared to other methods. As to Lung and Gisette datasets, CAUSE-FS achieves an average improvement of over 18\% in ACC and 23\% in NMI. As to Jaffe dataset, CAUSE-FS obtains an average improvement of over 13\% in both ACC and NMI. As to Umist and USPS datasets, CAUSE-FS gains over 9\% and 11\% average improvement in terms of ACC and NMI, respectively. As to COIL20 dataset, CAUSE-FS achieves almost 6\% average improvement in both ACC and NMI. Furthermore, to comprehensively assess the effectiveness of CAUSE-FS, we also present the results for all methods across varying numbers of selected features. Figs.~\ref{ACC} and~\ref{NMI} show the ACC and NMI values for all methods across different numbers of selected features. It is evident that CAUSE-FS outperforms other competing methods in most cases when the number of selected features ranges from 20 to 100. The superior performance of CAUSE-FS is attributed to two aspects: the joint learning of feature selection and confounder balancing, facilitating the identification of causally informative features, and the maintenance of a more reliable local structure by enhancing the importance of causal features in constructing the similarity graph.

\begin{figure*}[!htbp]
	\centering
	\includegraphics[width=\textwidth]{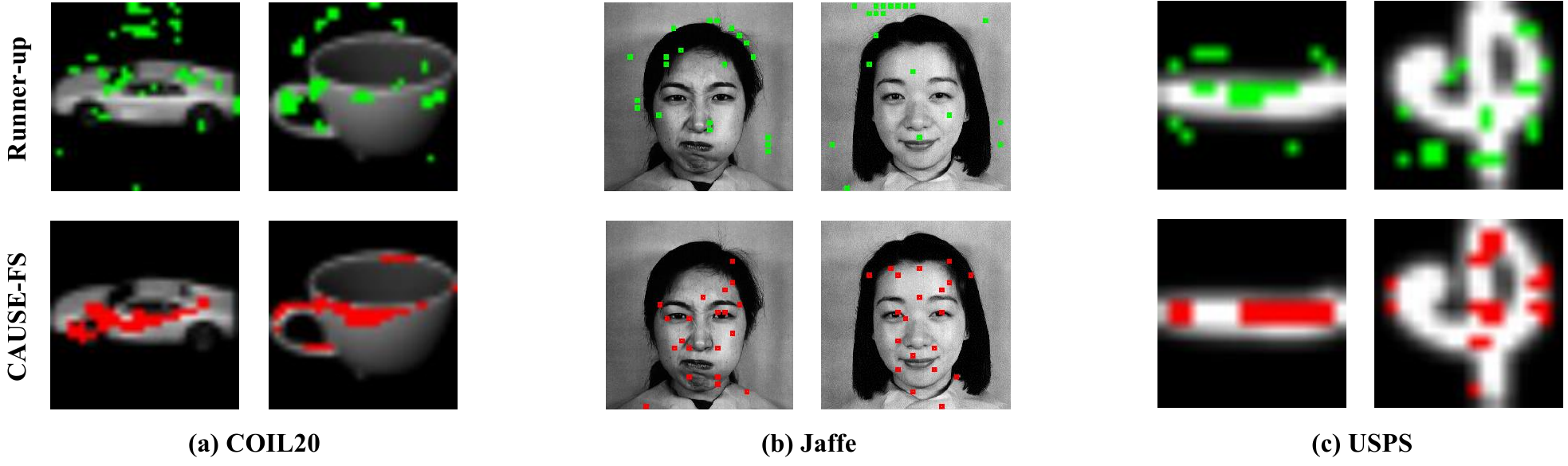} 
	\caption{Visualization of features selected by the runner-up method (the first row) and CAUSE-FS (the second row) on COIL20, Jaffe, and USPS datasets.}
	\label{Vis}
\end{figure*}

\subsubsection{Interpretability} To illustrate the interpretability of CAUSE-FS, we provide visualizations of the features selected by both CAUSE-FS and the runner-up method on  COIL20, Jaffe, and USPS datasets. Due to limited space, we only present a few examples in Fig. \ref{Vis}. As shown in Fig. \ref{Vis}, CAUSE-FS selects features concentrated in the critical regions of objects, such as the body of a car, a person's face, and the contours of a digit, whereas the runner-up methods tend to select numerous irrelevant contextual features. From the perspective of interpretability, CAUSE-FS can select causally informative features from unlabeled data to enhance downstream tasks like clustering. This ability enables it to explain why an image is assigned to the ``car'' cluster by identifying causal features, such as the car body.

\subsubsection{Parameter Sensitivity}
\begin{figure}[!htbp]
	\centering
	\includegraphics[width=\columnwidth]{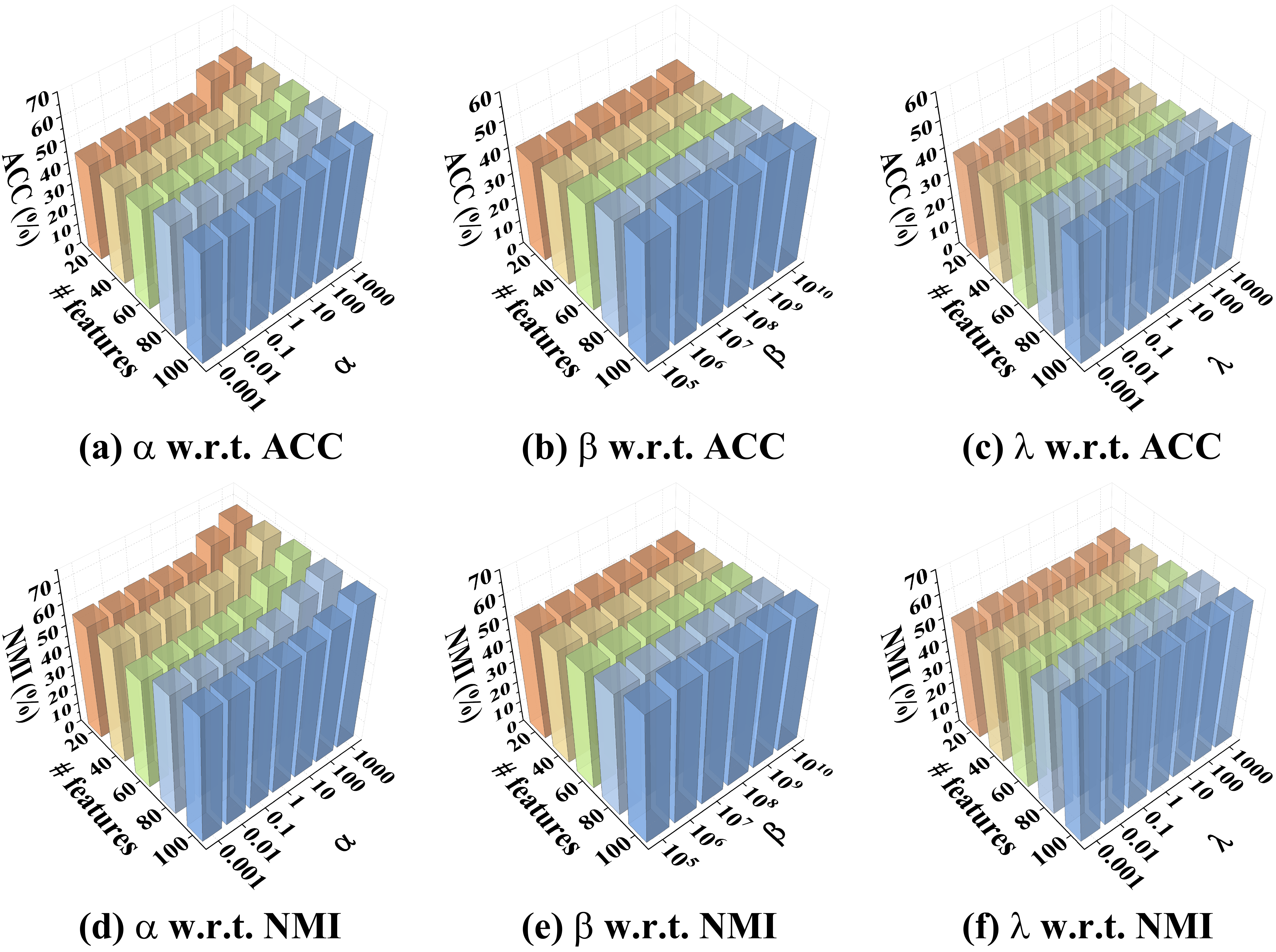} 
	\caption{ACC and NMI of CAUSE-FS with varying parameters $\alpha$, $\beta$, $\lambda$ and feature numbers on COIL20 dataset.}
	\label{Sensi}
\end{figure}

The proposed method CAUSE-FS involves three tuning parameters, i.e., $\alpha$, $\lambda$ and $\beta$. In this section, we investigate how the parameters in CAUSE-FS influence the performance of feature selection. Fig.~\ref{Sensi} shows the ACC and NMI results of CAUSE-FS as one parameter is varied while the others are kept fixed, across different numbers of selected features. We can observe that CAUSE-FS exhibits minor fluctuations when $\alpha$ is around 100, whereas it maintains relative stability w.r.t.  $\beta$ and $\lambda$.  

\begin{figure}[t]
	\centering
	\includegraphics[width=\columnwidth]{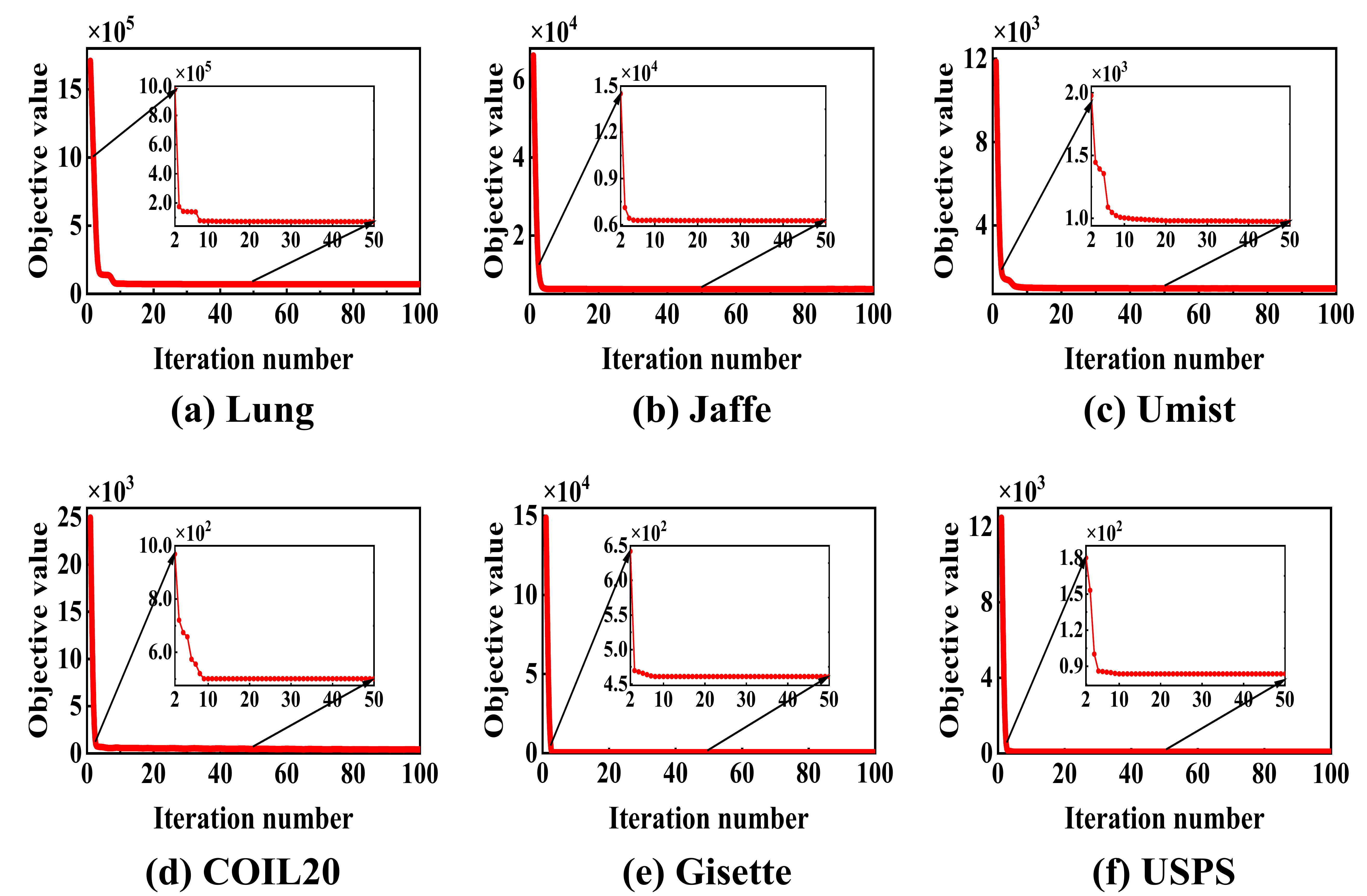} 
	\caption{Convergence curves of CAUSE-FS on on six datasets.}
	\label{Conver}
\end{figure}

\subsubsection{Convergence Analysis}
The convergence of Algorithm 1 was proved through theoretical analysis in Section~\ref{Convergence Analysis}. In this section, we experimentally investigate the convergence behavior of Algorithm 1. Fig.~\ref{Conver} displays the convergence curves of CAUSE-FS on six datasets, where the x-axis represents the number of iterations and the y-axis denotes the objective function value. From Fig.~\ref{Conver}, it can be observed that the convergence curves of CAUSE-FS exhibit a rapid decline within the first few iterations and stabilize after 20 iterations.

\subsubsection{Ablation Study}
We conduct an ablation study comparing CAUSE-FS with two of its variants to evaluate the contributions of its two key modules.

CAUSE-FS-\uppercase\expandafter{\romannumeral1} (without generalized causal regression module):
\begin{equation}\label{V1}
\begin{aligned}
&\min_{\bm{\Theta_{1}}}\frac{1}{2}\sum_{i,j=1}^{n}\sum_{m=1}^{M}\left[\|\bm{X}_{\cdot i}^{T}\bm{R}^{(m)}\bm{W}-\bm{X}_{\cdot j}^{T}\bm{R}^{(m)}\bm{W}\|_{2}^{2}\mathnormal{S}_{ij}^{(m)}\right.\\
&+\!\gamma_{m}\|\bm{S}^{(m)}\!\|_{\mathrm{F}}^{2}]\!-\!\!\operatorname{Tr}(\bm{G}^{\top}\!\!\sum_{m=1}^{M}\!\nu_{m}\bm{S}^{(m)})\!+\!\xi\|\bm{G}\|_{\mathrm{F}}^{2}\!+\!\lambda\|\bm{W}\|_{2,1},\\
\end{aligned}
\end{equation}
where $\bm{\Theta_{1}}=\{\bm{W},\bm{G},\{\bm{S}^{(m)}\}_{m=1}^{M}\}$.

CAUSE-FS-\uppercase\expandafter{\romannumeral2} (without multi-granular adaptive graph learning module):
\begin{equation}\label{V2}
\begin{aligned}
&\min_{\bm{\Theta_{2}}}\!\sum_{i=1}^{n}\alpha\|\mu_{i}\bm{X}_{\cdot i}^{\top}\bm{W} \!-\! \bm{F}_{i \cdot} \|_{\mathrm{2}}^{2}+\operatorname{Tr}( \bm{F}^{\top}\bm{L}_{S}\bm{F})+\lambda\|\bm{W}\|_{2,1}\\
&+\beta \sum_{r=1}^{d}\| \frac{1}{|\Lambda_{t}|}\sum_{i\in\Lambda_{t}}\phi(\mu_{i}Z_{\cdot i}^{r}) - \frac{1}{|\Lambda_{c}|}\sum_{j\in\Lambda_{c}}\phi(\mu_{j}Z_{\cdot j}^{r})\|_{\mathcal{H}_{\kappa}}^{2},\\
\end{aligned}
\end{equation}
where $\bm{\Theta_{2}}=\{\bm{W},\bm{F},\bm{\mu}\}$. The optimization constraints on the variables in CAUSE-FS-\uppercase\expandafter{\romannumeral1} and CAUSE-FS-\uppercase\expandafter{\romannumeral2} are identical to those in Eq.~(\ref{3.5}).

Table~\ref{Ablation} shows the results of the ablation experiments on six datasets. We observe that the performance of CAUSE-FS-\uppercase\expandafter{\romannumeral1} significantly decreases compared to CAUSE-FS regarding ACC and NMI. This highlights the effectiveness of the joint learning of feature selection and confounding balancing in identifying causally informative features. Moreover, CAUSE-FS consistently outperforms CAUSE-FS-\uppercase\expandafter{\romannumeral2} across all datasets, demonstrating that the multi-granular adaptive graph learning module can construct a more reliable similarity graph.

\begin{table}[!htbp]
	\centering
	% \normalsize
	\small
	\caption{\normalsize Average performance comparison of CAUSE-FS and its two variants in terms of ACC and NMI.}\label{Ablation}
	\setlength{\tabcolsep}{2.15mm}
	\begin{tabular}{lllllll}
		\toprule
		\multirow{2}*{Datasets} & \multicolumn{2}{c}{CAUSE-FS}&\multicolumn{2}{c}{CAUSE-FS-\uppercase\expandafter{\romannumeral1}} &\multicolumn{2}{c}{CAUSE-FS-\uppercase\expandafter{\romannumeral2}} \\
		\cmidrule(r){2-3} \cmidrule(r){4-5} \cmidrule(r){6-7}
		~&ACC&NMI&ACC&NMI&ACC&NMI\\
		% \hline
		\midrule
		Lung  &\textbf{81.43}  & \textbf{63.40} &44.99  &24.98   & 77.31 &58.94 \\
		Jaffe   &\textbf{85.08}  &\textbf{89.26}  &59.49  &61.90   &80.11  &83.57 \\
		Umist   &\textbf{55.99}  &\textbf{72.10}  &35.57  &46.38   &53.29  &66.78 \\
		COIL20   &\textbf{59.01}  &\textbf{70.00}  &33.79  &46.41   &56.72  &66.13 \\
		Gisette   &\textbf{83.32}  &\textbf{33.01}  &57.38  &2.86   &77.42  &22.86 \\
		USPS   &\textbf{66.82}  &\textbf{60.31}  &39.46  &26.44   &61.75  &50.82 \\
		\bottomrule
	\end{tabular}
\end{table}

\section{Conclusion}\label{Conclusion}
In this paper, we have proposed a novel UFS method from the causal perspective, addressing the limitations of existing UFS methods that overlook the underlying causal mechanisms within data. The proposed CAUSE-FS framework integrates feature selection with confounding balance into a joint learning model, effectively mitigating spurious correlations between features and clustering labels, thereby facilitating the identification of causal features. Furthermore, CAUSE-FS enhances the reliability of local data structures by prioritizing the influence of causal features over non-causal features during the construction of the similarity graph. Extensive experiments conducted on six benchmark datasets demonstrate the superiority of CAUSE-FS over SOTA methods.

\end{document}